# A Novel Method for Automatic Segmentation of Brain Tumors in MRI Images

Saeid Fazli
Research Institute of Modern Biological Techniques
University of zanjan
Zanjan, Iran
fazli@znu.ac.ir

Parisa Nadirkhanlou
Electrical Engineering Department
University of zanjan
Zanjan, Iran
nadirkhanlou@znu.ac.ir

*Abstract*—The brain tumor segmentation on MRI images is a very difficult and important task which is used in surgical and medical planning and assessments. If experts do the segmentation manually with their own medical knowledge, it will be time-consuming. Therefore, researchers propose methods and systems which can do the segmentation automatically and without any interference. In this article, an unsupervised automatic method for brain tumor segmentation on MRI images is presented. In this method, at first in the pre-processing level, the extra parts which are outside the skull and don't have any helpful information are removed and then anisotropic diffusion filter with 8-connected neighborhood is applied to the MRI images to remove noise. By applying the fast bounding box(FBB) algorithm, the tumor area is displayed on the MRI image with a bounding box and the central part is selected as sample points for training of a One Class SVM classifier. A database is also provided by the Zanjan MRI Center. The MRI images are related to 10 patients who have brain tumor. 100 T2-weighted MRI images are used in this study. Experimental results show the high precision and dependability of the proposed algorithm. The results are also highly helpful for specialists and radiologists to easily estimate the size and position of a tumor.

I. INTRODUCTION

By the increasing age population, cancer has become a worldwide health problem. In accordance to the most recent figures of World Cancer Research Fund, the first reason of death in world is cancer [1]. Nobody knows the exact reason of brain tumors creation. Doctors can barely explain about how someone is overtaken by brain tumor and someone else isn't.

While most of the natural cells are getting old or damaged, they disappear and new cells are replaced with them. Sometimes, this process goes wrong; New cells are produced when body doesn't need them and the old and damaged cells don't disappear. Therefore, the illimitable and incontrollable increase of cells causes the brain tumor creation. If the brain tumors are not diagnosed immediately, they could either cause a serious brain damage or even death.

In all of treatment methods, any information about position and size of the tumor for successful treatment is essential [2]. Awareness of the tumor position and size, especially changes about tumor size, can provide very important information to find the most effective regime for the patients during the treatment, including surgery, radiotherapy and chemotherapy [2].

Recently MRI imaging technique is taken into consideration. MRI provides a view from inside human body. High spatial resolution and excellent soft tissue diagnosis are the advantages of MRI over other medical imaging techniques. Also there are no means to entrance and no drug for injection into the human body in MRI imaging process. The entire process has not any radiation damage and is completely safe [3]. Also Computed Tomography (CT), Positron Emission Tomography (PET), CT/PET are other techniques for medical imaging.

The traditional explanation of MRI images by a proficient is a very difficult and time-consuming task. Also the result directly depends on the experience of the proficient. Accordingly, finding an accurate and fully automatic method to provide the information to the doctors is consequential [2].

There are many proposed techniques for automatic and semi-automatic detection and segmentation of brain tumors. The proposed techniques can be mainly divided into two groups; Intelligent based and non-intelligent based. Most of the leading intelligent based systems are artificial neural networks [4], fuzzy c-means (FCM) [5,6,7], fuzzy connectedness [2], support vector machine (SVM) [8,9], particle swarm optimization (PSO) [10], genetic algorithm [11] and hybrid methods. On the other hand, the leading non-intelligent techniques include thresholding [12,13] and region growing [14,15] and etc. Usually the combination of these algorithms are used to achieve better results [16,17].

Purpose of image segmentation is to segment an image into regions which are meaningful for a particular task. So its a widespread field not only in medical imaging, but also in computer vision and satellite imagery. The choice of a particular method among various methods and approaches are used, depends on the characteristics of the case [18].

There are many problems and challenges about brain tumors segmentation. Brain tumors may appear in any size, shape and location.

One challenge in tumor detection and segmentation is how to investigate the real data's nonlinear distribution issue. To decide whether some data are in target class or not is the goal of the one-class classification [9]. Owing to its ability of learning the nonlinear distribution of the real data without using any prior knowledge, one-class support vector machines (SVMs) have been applied in tumor segmentation [8].

The rest of this paper is organized as follow: Section II explains the methods and materials. Section III is dedicated to the proposed approach. The experimental results are presented in section IV. Finally, conclusions forms the last section.

## II. METHODS AND MATERIALS

### A. Pre-processing

The pre-processing phase has a great importance in the applications of image processing and specially segmentation. Generally in the pre-processing phase, the main goal is to remove the noise from the images. Undoubtedly MRI images have noises which have to be removed. But the noise deletion shouldn't destroy the edges of the image and decrease the clarity and quality of it. There are several methods for removing noise, including: Gaussian filter [19], contourlet transform approach and wavelet thresholding approach [20], median filter [6], anisotropic diffusion filter [2].

Anisotropic diffusion filter is a method for removing noise which is proposed by Persona and Malik [21]. This method is for smoothing the image by preserving needed edges and structures. Fundamental idea is to adjust the smoothing level in a region based on the edge structure in the neighborhood. Homogenous regions are highly smoothed and strong edge regions are barely smoothed (to preserve the structure)

### B. Fast Bounding Boxes algorithm

In each input MRI slice (axial view), there is a left–right axis of symmetry of the brain. A tumor which is considered an abnormality in the brain, typically perturbs this symmetry. Thus an axis-parallel rectangle on the right side that is very dissimilar from its reflection about the axis of symmetry on the left side—i.e., the gray level intensity histograms of the inside of the two rectangles are most dissimilar and the outside of the rectangles are relatively similar. A novel score function utilize that can identify the region of change with two very rapid searches along the vertical and horizontal direction of the image. Bhattacharya coefficient (BC) measures similarity between two normalized gray level intensity histograms. When two normalized histograms are the same, the BC between them is 1 and when two normalized histograms are completely dissimilar, the associated BC value is 0 [22].

### C. Support Vector Machine (SVM)

Support Vector Machines is a special family of learning machines that were first proposed by Vapnik. SVM is a classification algorithm for analyzing high-dimensional data, and has the ability to learn the nonlinear distribution of the real data without using any prior knowledge [3]. As in the Statistical Learning Theory, optimum efficiency of SVM can be obtained by forming the optimum classification level with the greatest classification margin [23].

One-class SVM forms a classifier just from a collection of labeled positive templates called "positive training samples" [24]. Suppose that the user has the sequent training data set $X=\{x_i \quad i=1,2,3,...\ell\}$. Where $x_i$ is the i th observation and $\ell \in N$ is the number of observation. Presume that there is a feature map which maps the training data into a higher-dimensional internal product space called feature space F, i.e.

$$\phi: X \to F \qquad (1)$$

Training sample $x_i$ in $X \to \phi(x_i)$ in F.

If there is a function $f$ which takes the amount +1 for tumor data and −1 for non-tumor data, after that in F, the data can be divided from the source with the maximal margin. So only the tumor data is considered and the object function is formulated as

$$\min_{W \in F, \eta \in R^\ell, b \in R} \frac{1}{2}W^T W + \frac{1}{vl}\sum_i \eta_i - b, \qquad (2)$$
$$s.t \quad W.\phi(x_i) \geq b - \eta_i, \eta_i \geq 0$$

W = the normal vector of hyperplane which represent the decision boundary.

b = represents the threshold of function $f$.

$\eta_i$ = the slack variable which is penalized in the objective function.

ν = regularization term, a user-defined parameter which controls the trade off and indicates the fraction of samples that should be accepted by the description.

Proper W and b are to be found to minimize (2). Here for each of the inequality constrains in (2), the positive Lagrange coefficients, $\alpha_i$ and $\beta_i$ (for i=1,2,3,...$\ell$), was introduced. This gives the following Lagrange form

$$L(W,\eta,b,\alpha,\beta) = \frac{1}{2}W^T W + \frac{1}{vl}\sum_i \eta_i - b$$
$$- \sum_i \beta_i \eta_i - \sum_i \alpha_i(W.\phi(x_i) - b + \eta_i) \qquad (3)$$

Where η, α and β are one-column vectors displaying [$\eta_i$], [$\alpha_i$] and [$\beta_i$], respectively. To minimize (3), let its gradient, with respect to W, b and $\eta_i$, individually, equal to zero, that is

$$\frac{\partial l}{\partial W} = W - \sum_i \alpha_i \phi(x_i) = 0 \Rightarrow W = \sum_i \alpha_i \phi(x_i) \quad (4)$$

$$\frac{\partial l}{\partial b} = -1 + \sum_i \alpha_i = 0 \Rightarrow \sum_i \alpha_i = 1 \qquad (5)$$

And
$$\frac{\partial l}{\partial \eta_i} = \frac{1}{vl} + \alpha_i - \beta_i = 0 \Rightarrow \alpha_i = \frac{1}{vl} - \beta_i \leq \frac{1}{vl} \quad (6)$$

Replacing (4)-(6) into (3), have been got

$$\min \quad \frac{1}{2}\sum_{i,j} \alpha_i \alpha_j k(x_i, x_j)$$
$$s.t. \quad 0 \leq \alpha_i \leq \frac{1}{vl}, \sum_i \alpha_i = 1 \quad (7)$$
$$k(x_i, x_j) = \phi(x_i).\phi(x_j)$$

Equation (7) can be further written in a more compressed matrix form

$$\min \quad \frac{1}{2}\sum_{i,j} \alpha^T Q \alpha$$
$$s.t. \quad 0 \leq \alpha_i \leq \frac{1}{vl}, e^T \alpha = 1 \quad (8)$$
$$\phi_{i,j} = k(x_i, x_j)$$

e = a measure vector of length N.

The dual problem in (8) shows a well-known quadratic form and its minimization can be solved by using the well-known quadratic programming (QP) optimization method. The optimal amount of α corresponds to the minimum of the objective function. Those objects with weight $\alpha_i$>0 are required in the final statement of the data set. They are commonly called support vectors in machine learning research. The optimal amount of the parameter b can be calculated by the (9)

$$b = \sum_j \alpha_j k(x_j, x_i) \quad (9)$$

$x_i$ = any one of the support vectors.

The tumor data can be classified, when the optimal values of the parameters are obtained, according to the following decision function

$$f(x) = \text{sgn}(\sum_i \alpha_i k(x_i, x) - b) \quad (10)$$

The data corresponding to $f(x) \geq 0$ are determined as tumor data candidates. If not, they are regarded as non-tumor candidates.

The learning ability of one-class SVM emanates from the "kernel trick" [24]. This trick is performed by different choice of k(x,y) introduced in (7). Notice that in the formulation of one-class SVM, the mapping φ is only assigned implicitly by kernel k(x,y). So there is no compulsion to give a clear mapping and the kernel is described instead.

An acceptable kernel should be described, i.e., an proper kernel can map the target data into a bounded spherically shaped area in the feature space and outline the objects outside the data boundary. With the "kernel trick", one-class SVM can deal with nonlinear multimode data distribution [8].

III. PROPOSED METHOD

In this article, a novel approach is presented for brain tumor segmentation on MRI images which is fully automatic and does not need any user interaction. Fig. 1 shows the block diagram of the proposed algorithm.

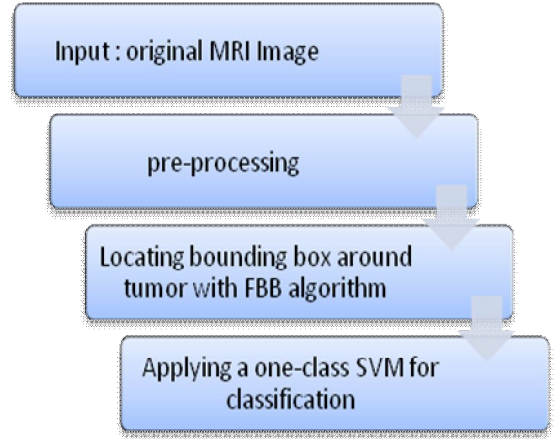

Figure 1. Block diagram of the proposed approach

The pre-processing phase has two steps:

1) In the first step the extra and useless parts outside the skull are removed. For this task, at first the boundary of the skull is determined by automatic global thresholding [25], then with creating a binary image which is head mask indeed, the extra regions from outside of the skull are removed. With this operation, the required calculations in later steps and total time of segmentation are decreased.

2) This step is to remove the noise. MRI images include image and some noise. This noises cause some disorders in image. They should be removed for segmentation process improvement without destroying the edges of the image and decreasing its clarity. Here, anisotropic diffusion filter with 8-connected neighborhood is applied on the image for removing noise. The essential idea of this approach is quite simple [26]:

$$I(x, y, t) = I_0(x, y) * G(x, y, t) \quad (11)$$

$I(x, y, t)$ = filtered image;

$I_0(x, y)$ = original image;

$G(x, y, t)$ = Gaussian kernel with different scale parameter t;

In this method, filtering can be reached as the iterative way of the heat diffusion equalization where $I(x,y,0) = I_0(x,y)$ is the primary condition, and is equivalent to time $t$. Perona and Malik introduced a solution for the previously stated problem, that permits space variant (anisotropic) blurring in order to locate edges with more accuracy. They used 4-connected neighborhood in their calculations [21]. But in this article the calculations related to 8-connected neighborhood are used.

The resulting filter is based on the following equation

$$I_{i,j}^{t+1} = I_{i,j}^t + \lambda(c_U^t \nabla_U I_{i,j}^t + c_D^t \nabla_D I_{i,j}^t \\ + c_L^t \nabla_L I_{i,j}^t + c_R^t \nabla_R I_{i,j}^t + \\ c_{UL}^t \nabla_{UL} I_{i-1,j-1}^t + c_{UR}^t \nabla_{UR} I_{i-1,j+1}^t \\ + c_{DL}^t \nabla_{DL} I_{i+1,j-1}^t + c_{DR}^t \nabla_{DR} I_{i+1,j+1}^t) \quad (12)$$

Where $I_{i,j}^t$ is the Pixel value at position (i,j) at the time t, the coefficient $\lambda \in [0, 0.25]$, and terms $\nabla$ are the finite differences calculated on a 8-connected neighborhood.

$$\nabla_U I_{i,j}^t = I_{i-1,j}^t - I_{i,j}^t \; ; \nabla_D I_{i,j}^t = I_{i+1,j}^t - I_{i,j}^t \\ \nabla_L I_{i,j}^t = I_{i,j-1}^t - I_{i,j}^t \; ; \nabla_R I_{i,j}^t = I_{i,j+1}^t - I_{i,j}^t \\ \nabla_{UR} I_{i,j}^t = I_{i-1,j+1}^t - I_{i,j}^t \; ; \nabla_{UL} I_{i,j}^t = I_{i-1,j-1}^t - I_{i,j}^t \\ \nabla_{DL} I_{i,j}^t = I_{i+1,j-1}^t - I_{i,j}^t \; ; \nabla_{UL} I_{i,j}^t = I_{i+1,j+1}^t - I_{i,j}^t \quad (13)$$

$c_U^t$, $c_D^t$, $c_L^t$, $c_R^t$, $c_{DL}^t$, $c_{DR}^t$, $c_{UL}^t$, $c_{UR}^t$ are also called conduction coefficients calculated at the same time t on the basis of diffusion function whose arguments are finite differences written above. Perona and Malik in [21] offer two diffusion functions to calculate the conduction coefficients.

$$g(\nabla I) = \exp(-\frac{\|\nabla I\|}{k})^2 \\ g(\nabla I) = 1/(1 + (-\frac{\|\nabla I\|}{k})^2) \quad (14)$$

k= a constant value [13];

Brain has a left-right natural symmetry and with appearance of tumor, this symmetry is disarranged. In FBB approach with considering similarity and dissimilarity of gray level intensity histogram of symmetrical regions, then Bhattacharya coefficient is calculated and the region of tumor is automatically marked by a bounding box.

In this method one class SVM is chosen as classifier and tumor pixels is used as training set. Since tumors are detected in different shapes the parts that is extracted by bounding box may include pixels of healthy part of brain in order to overcome this drawback, just the central part is selected as sample points. Here, the radius basis function (RBF) was chosen as the learning kernel.

Another point to be taken in to account, is that SVM simplifies well in high-dimensional spaces and feature extraction can be accomplished in the training step of SVM. Consequently no special feature extraction technique is needed in this approach.

## IV. EXPERIMENTAL RESULTS

The required database for this study is provided by the Zanjan MRI Center. The MRI images are related to 10 patients who have brain tumor and a total number of 100 T2-weighted MRI images is used. Some of the results of applying the proposed algorithm are shown in Fig. 2. The first row presents original brain MRI images, the second row are the results of the pre-processing and FBB, and the third row are the final results of a one class SVM classification.

Various outcomes of the classification is given in Table 1.

Table 1. Possible outcomes

| Real Group | Segmentation Result | |
|---|---|---|
| | Tumor | Non-tumor |
| Tumor | TP | FN |
| Non-tumor | FP | TN |

Two criteria including accuracy as in [10] and similarity index (SI) as in [27] are calculated for evaluation of the proposed method. Then the results are compared to the existing methods. Outcome of the comparison are presented in Table 2.

$$Accuracy = \frac{TP+TN}{TP+TN+FP+FN} \\ SI = \frac{2TP}{2TP+FP+FN} \quad (15)$$

As can be seen in Table 2, each criterion has three columns. The first column is the results of the proposed approach. The second column shows the results of applying anisotropic diffusion filter with 4-connected neighborhood in the per-processing phase of the presented approach. The third column is the results of the approach presented in [13]. In the last row the mean results of each method are calculated.

## V. CONCLUSION

In this article, a fully automatic method for tumor segmentation on MRI images is presented. This method has three main steps: The First step is a pre-processing task in which the extra and useless parts of skull are removed and also anisotropic diffusion filter is applied to the image by 8-connected neighborhood for removing noise from it. In the second step, using FBB, the required training set for One Class SVM is obtained automatically. In the last stage, using a One Class SVM classifier with RBF kernel, the tumor position is separated from the healthy textures. By comparing the results of the proposed approach to the existing approaches which are shown in table 2, it can be clearly seen that the proposed approach is more reliable and accurate.

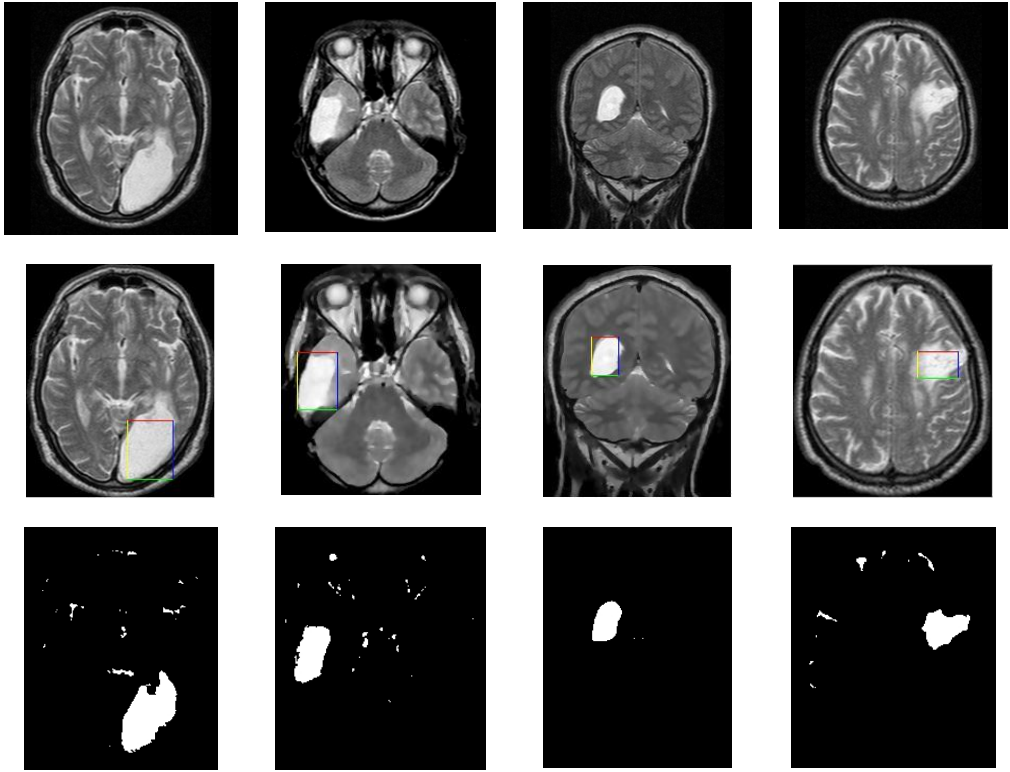

Figure 2. The first row shows the original brain MRI images, the second row are the results of the pre-processing and FBB, and the third row are the final results of one class SVM classification

Table 2. The first column of each criterion presents the results of the proposed approach, The second column of each criterion are the results of the applied anisotropic diffusion filter with 4-connected neighborhood in per-processing phase of the presented approach. The third column of each criterion presents outcome of the presented approach in [14]. In the last row the mean results of each method are calculated.

|  | Accuracy | | | SI | | |
|---|---|---|---|---|---|---|
| **Image set 1** | 98.24% | 98.12% | 97.02% | 0.8973 | 0.8530 | 0.7945 |
| **Image set 2** | 96.67% | 96.12% | 96.01% | 0.8533 | 0.7983 | 0.7712 |
| **Image set 3** | 98.50% | 98.27% | 93.26% | 0.9141 | 0.8645 | 0.6338 |
| **Image set 4** | 98.00% | 97.10% | 96.79% | 0.8500 | 0.7940 | 0.6503 |
| **Image set 5** | 99.01% | 98.84% | 98.35% | 0.9865 | 0.9483 | 0.8116 |
| **Image set 6** | 98.14% | 97.97% | 95.85% | 0.8704 | 0.8300 | 0.6413 |
| **Image set 7** | 97.39% | 97.41% | 97.48% | 0.8697 | 0.8396 | 0.8051 |
| **Image set 8** | 99.29% | 99.20% | 98.23% | 0.8824 | 0.8414 | 0.8100 |
| **Image set 9** | 97.45% | 96.87% | 94.19% | 0.8792 | 0.8368 | 0.6400 |
| **Image set 10** | 97.43% | 96.77% | 92.00% | 0.8750 | 0.8205 | 0.5559 |
| **Total mean** | 98.01% | 97.67% | 95.92% | 0.8878 | 0.8426 | 0.7114 |